# Adopting Trustworthy AI for Sleep Disorder Prediction: Deep Time Series Analysis with Temporal Attention Mechanism and Counterfactual Explanations

Pegah Ahadian, Kent State University
Wei Xu, Brookhaven National Laboratory
Sherry Wang, Chapman University
Qiang Guan, Kent State University
pahadian@kent.edu
qguan@kent.edu

**Abstract**

Sleep disorders have a major impact on both lifestyle and health. Effective sleep disorder prediction from lifestyle and physiological data can provide essential details for early intervention. This research utilizes three deep time series models and facilitates them with explainability approaches for sleep disorder prediction. Specifically, our approach adopts Temporal Convolutional Networks (TCN), Long Short-Term Memory (LSTM) for time series data analysis, and Temporal Fusion Transformer model (TFT). Meanwhile, the temporal attention mechanism and counterfactual explanation with SHapley Additive exPlanations (SHAP) approach are employed to ensure dependable, accurate, and interpretable predictions. Finally, using a large dataset of sleep health measures, our evaluation demonstrates the effect of our method in predicting sleep disorders.

**Introduction**

The way complex data is analyzed and processed has changed dramatically in recent years due to the integration of Artificial Intelligence (AI) into a variety of fields [1]. Among its multiple applications, artificial intelligence (AI) has significantly improved healthcare by making it possible to develop predictive models that are highly accurate at both diagnosing and predicting a wide range of medical disorders. Millions of people worldwide struggle with sleep difficulties, which is one of the disorders that AI-driven predictive analytics can help with. According to Han et al. (2012) [2], stress, sleep apnea, insomnia, and other sleep disorders can cause serious health problems such as depression, diabetes, and heart disease. To reduce these risks and enhance the general quality of life for those who are impacted, early detection and intervention are essential.

Despite the promising capabilities of AI in healthcare, one of the primary challenges hindering its widespread adoption is the "black-box" nature of many AI models. These models, especially deep learning architectures, often work with a high level of complexity and opacity, making it difficult for users to understand the underlying decision-making processes. This lack of transparency can lead to distrust and reluctance in relying on AI based systems, especially in critical fields like healthcare where decisions have significant consequences.

Explainable AI (XAI) [3] [4] is a significant research area to address this problem. The goal of XAI is to improve the understanding and transparency of how AI models work, thereby making it more trustworthy to integrate them into real-world applications. When applying time series analysis for the purpose of predicting sleep disorders, XAI techniques provide important insights into the ways in which particular variables affect the predictions made by AI models. This transparency is essential for healthcare professionals to validate and build confidence in the AI-driven diagnoses and recommendations.

Our paper presents a Trustworthy AI approach with Explainable Time Series Analysis (XTSA), which aims to enhance trust in AI-driven sleep problem predictions and provide practical insights. First, existing deep learning models that excel at analyzing sequential data are incorporated into our work, such as Temporal Convolutional Networks (TCN) and Long Short-Term Memory (LSTM) networks, and Temporal Fusion Transformer (TFT). Then we employ counterfactual explanations to provide an enhanced comprehension of the model's decision-making process by showing how minor modifications to input features can affect the predictions. Our XTSA system combines robust deep learning models with counterfactual explanations using SHAP (SHapley Additive exPlanations) values and temporal attention mechanism.

**Motivation**

**Enhanced Interpretability**

Understanding the complex interplay of demographic, physiological, and behavioral factors over time is crucial for predicting sleep disorders. These disorders are influenced by how these factors interact and evolve, making the temporal dynamics essential for accurate assessments and interventions.

Our approach leverages SHAP values integrated with time series models like LSTM and TCN to enhance interpretability. SHAP values clarify how each input feature, such as stress levels, sleep duration, and quality of sleep, affects predictions at various time steps, helping healthcare professionals identify key periods for potential intervention.

Moreover, time series models capture complex patterns and dependencies that traditional models may overlook, such as sudden changes in stress or consistent declines in sleep quality, which could signal the onset of sleep disorders. These insights enable practitioners to better understand and respond to the dynamics of sleep disorders, aiding in the creation of personalized treatment plans that consider the unique factors influencing each patient.

**The Necessity of Temporal Analysis**

Sleep disorders are inherently time-dependent conditions, with factors such as stress levels, physical activity, and sleep quality fluctuating significantly over time. These fluctuations can have a profound impact on sleep health, necessitating the use of models that can effectively analyze temporal data. Time series models like Long Short-Term Memory (LSTM) networks, Temporal Convolutional Networks (TCN), and Temporal Fusion Transformer (TFT) are particularly well-suited for this task.

LSTM and TFT networks are capable of capturing long-term dependencies in sequential data, which is crucial for understanding how past behaviors and conditions affect current sleep health. By learning these dependencies, LSTM models provide insights into how historical data points influence current predictions, offering a deeper understanding of the temporal dynamics involved in sleep disorders.

TCNs are designed to capture temporal patterns and trends over varying time scales, making them ideal for analyzing the progression of sleep disorders. This capability helps in understanding how certain behaviors or conditions evolve over time and impact sleep health.

Moreover, time series models like LSTM, TCN, TFT can handle irregular intervals between data points, a common occurrence in real-world healthcare data where measurements may not be taken consistently. This flexibility ensures that the models can effectively process and analyze data even when collected at varying frequencies, making them robust tools for predicting sleep disorders in diverse clinical scenarios.

By focusing on these aspects, our work provides a comprehensive approach to understanding and predicting sleep disorders, ultimately leading to more effective interventions and better patient outcomes.

**Related Work**

Time series analysis within the healthcare sector has greatly benefited from the integration of deep learning techniques. Notably, Long Short-Term Memory (LSTM) networks and Temporal Convolutional Networks (TCNs) have become instrumental due to their proficiency in capturing temporal dependencies essential for analyzing health data. LSTM networks, pioneered by Hochreiter and Schmidhuber (1997), excel in handling long-term dependencies, making them particularly effective for predicting health outcomes over extended periods. For instance, Sano et al. (2018) utilized LSTMs to predict sleep disorders by analyzing physiological and lifestyle data, highlighting the model's utility in chronic health condition monitoring [5].

Contrastingly, TCNs leverage convolutional layers to process sequential data, offering a computationally efficient alternative to recurrent networks. This model's ability to manage long-range dependencies with reduced computational overhead has proven beneficial in applications such as sleep staging and arrhythmia detection [6]. The Temporal Fusion Transformer (TFT) model further exemplifies the evolution of time series models, providing interpretable multi-horizon forecasting capabilities that enhance the traditional models' transparency, crucial for clinical decision-making [7].

The advancement of Explainable AI (XAI) has addressed the opaque nature of traditional deep learning models, fostering trust and reliability in healthcare applications. Techniques like SHAP (SHapley Additive exPlanations) have been particularly transformative, offering insights into how various features influence model predictions [8]. These techniques are critical in contexts where understanding model behavior can directly impact clinical outcomes.

For example, Zogan et al. (2022) demonstrated the potential of XAI in mental health by using a hybrid deep learning model for depression detection, where explainability played a crucial role in interpreting the influence of multi-aspect social media features on model predictions [9]. Similarly, research by Vaquerizo-Villar et al. (2023) compared different deep learning architectures for sleep staging in pediatric patients, showing how XAI could enhance model transparency and facilitate clinical decision-making [10].

The fusion of deep learning with XAI not only augments model accuracy but also ensures the interpretability necessary for healthcare professionals to adopt these technologies confidently. Studies have consistently shown that enhancing model transparency directly correlates with improved patient care outcomes, underlining the importance of XAI in operational healthcare settings [11].

Our study contributes to this evolving landscape by integrating advanced time series models with robust explainability frameworks to predict sleep disorders, thus providing actionable insights that enhance clinical trust and patient care. This synergy of time series analysis and XAI not only advances our understanding of sleep health dynamics but also sets a benchmark for future research in the domain.

**Research Methods**

This study employs advanced deep learning models Long Short-Term Memory (LSTM) networks, Temporal Convolutional Networks (TCN), and Temporal Fusion Transformer (TFT) to efficiently analyze sequential data for predicting sleep disorders. Each model is chosen for its specific capabilities in capturing temporal dependencies crucial for time series analysis.

LSTM networks are a type of recurrent neural network designed to process sequential data by maintaining information across time steps through its unique architecture comprising input, forget, and output gates [12].

These gates are mathematically expressed as:

$$i_t = \sigma(W_i \cdot [h_{t-1}, x_t] + b_i)$$
$$f_t = \sigma(W_f \cdot [h_{t-1}, x_t] + b_f)$$
$$o_t = \sigma(W_o \cdot [h_{t-1}, x_t] + b_o)$$
$$C_t = f_t * C_{t-1} + i_t * \tanh(W_C \cdot [h_{t-1}, x_t] + b_C)$$
$$h_t = o_t * \tanh(C_t)$$

where $i_t$, and $f_t$ represent the input, forget, and output gates, respectively. $c_t$ is the cell state, $h_t$ is the hidden state, and $x_t$ is the input at time step t. The LSTM model processes sequential data by maintaining the cell state $c_t$ across time steps, allowing it to capture long-term dependencies that are crucial for time series forecasting.

TCNs utilize dilated convolutional layers to capture temporal patterns without the need for recurrent connections, offering an efficient alternative for parallel processing [13].

The output of a TCN is computed as:

$$h_t = \text{ReLU}(W_d * x_t + b)$$

where $W_d$ represents the convolutional filter with dilation d, $x_t$ is the input sequence, and b is the bias term. By adjusting the dilation rate dd, TCNs can model long-range dependencies across different time scales without requiring recurrent connections, making them well-suited for parallel processing and efficient learning.

TFT combines the benefits of attention mechanisms and recurrent neural network architectures, optimizing it for complex forecasting tasks involving time series data [7].

$$V_t = W_v \cdot x_t + b_v$$

where $W_v$ and $b_v$ are learnable weights and biases, and $V_t$ represents the selected variable set at time t.

To enhance model interpretability, we incorporate SHapley Additive exPlanations (SHAP), which quantify each feature's contribution to the prediction, providing a transparent and understandable model output. Additionally, Temporal Attention mechanisms are applied within our LSTM model which the attention score $a_t$ for time step $t$ is calculated as:

$$a_t = \frac{\sum_{t'} \exp(e_{t'})}{\exp(e_t)}$$

where $e_t$ is the score computed by an alignment function (e.g., a feedforward network) based on the current hidden state and the input at time step t. The final context vector $c_t$ is then obtained by taking a weighted sum of the hidden states:

$$c_t = \sum_t a_t h_t$$

This context vector $c_t$ captures the most relevant information across the time steps, allowing the model to make more informed predictions.

Counterfactual Explanations are also used to demonstrate how altering input features could change predictions, providing insights into the model's decision-making process and helping to identify key determinants of sleep disorders.

By integrating these deep learning and interpretability techniques, our approach aims to deliver both accurate and transparent predictions, addressing the need for trustworthiness in AI applications for healthcare.

**Dataset**

The dataset employed in this study originates from the "Multilevel Monitoring of Activity and Sleep in Healthy People" [14]. Due to the absence of directly usable features, we derived several key attributes encompassing demographic information, lifestyle habits, and physiological measurements. This section details the dataset structure and elaborates on the sleep disorders analyzed.

The dataset incorporates diverse features critical for predicting sleep disorders. Demographic features include age, gender, and socioeconomic indicators, which influence sleep patterns and disorder prevalence. Lifestyle attributes, such as stress level (measured on a scale of 1-10), physical activity (normalized to 30-90 minutes daily), and quality of sleep (rated 1-10), provide insights into behavioral impacts on sleep health. Physiological metrics, like BMI (calculated using standard formulas) and average heart rate, capture underlying health conditions. Preprocessing addressed missing data through imputation, normalized numerical values, and applied feature scaling to ensure consistency, improving model accuracy and reliability.

We normalized physical activity levels calculated from start and end times of activities, adjusting the total activity time to a standard scale ranging from 30 to 90 minutes. This normalization provided a consistent metric for physical activity across all subjects.

Sleep duration was derived from total sleep times and converted into hours. We also integrated cardiovascular health indicators by averaging multiple heart rate readings per subject.

The training set comprised data from 400 patients with 14 distinct features, which data augmentation employed to extend the data for the training step, while the test dataset included 22 patients. Key metrics recorded included:

Sleep Duration: Averaged 7.13 hours, with a range from 5.8 to 8.5 hours.

Quality of Sleep: Assessed on a scale from 1 to 10, with an average score of 7.31 and a range from 4 to 9.

Physical Activity Level: Averaged 59.17 minutes per day, spanning from 30 to 90 minutes.

   Stress Level: Scored on a scale from 1 to 10, averaging 5.39 with a range from 3 to 8.

The dataset included indicators for the presence or absence of sleep disorders such as Insomnia and Sleep Apnea.

**Experimental Result**

In our evaluations, we focused on the performance of three distinct models: LSTM, TCN, and TFT. These models were chosen due to their relevance in handling time-series data, which is critical for capturing temporal dependencies in sleep health datasets.

The LSTM model achieved a training accuracy of 93.72\%, with a validation accuracy initially reported as 90.00\%. After reviewing the experimental data on TCN, we confirmed that training accuracy is indeed 90.00\%, aligning with the testing accuracy of 85.62\%. The validation loss was noted at 0.4958, indicating effective performance on unseen data.

Contrary to an earlier misreport, the TFT model's validation accuracy was accurately recorded at 89.33\%, which is comparably higher than the TCN model but slightly less than the initially misstated LSTM's performance in Table 1. The validation loss for the TFT was also favorably low, demonstrating its strong generalization capabilities across different test scenarios.

### TABLE I: Model Performance Comparison

| Model | Training Accuracy (%) | Testing Accuracy (%) |
| --- | --- | --- |
| LSTM | 93.72 | 90.00 |
| TCN | 90 | 85.62 |
| TFT | 91 | 89.33 |

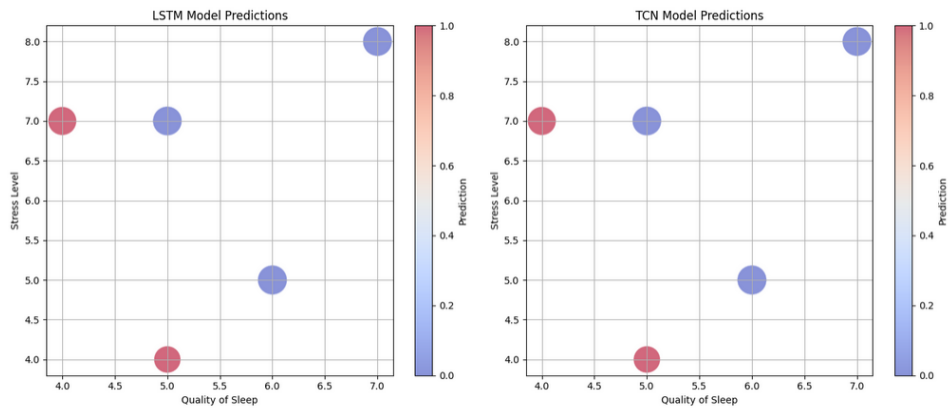

*Fig. 1 Model Predictions Based on Quality of Sleep, Stress Level, and Heart Rate. Color of points: Predictions made by the LSTM/TCN model (0 for no disorder, 1 for disorder)*

This case study highlights the robustness of LSTM, TCN, and TFT models in handling specific scenarios within the healthcare domain. The results reinforce the importance of model interpretability and accuracy in clinical settings, ensuring healthcare professionals can trust AI-driven predictions for better patient outcomes. Both models made consistent predictions for patients with Quality of Sleep scores of 5 or less combined with high Stress Level and Heart Rate values, predicting the presence of a sleep disorder. For patients with higher Quality of Sleep scores and relatively lower stress or heart rate values, the models predicted the absence of a sleep disorder. Figure 1 shows a scatter plot of LSTM and TCN model predictions.

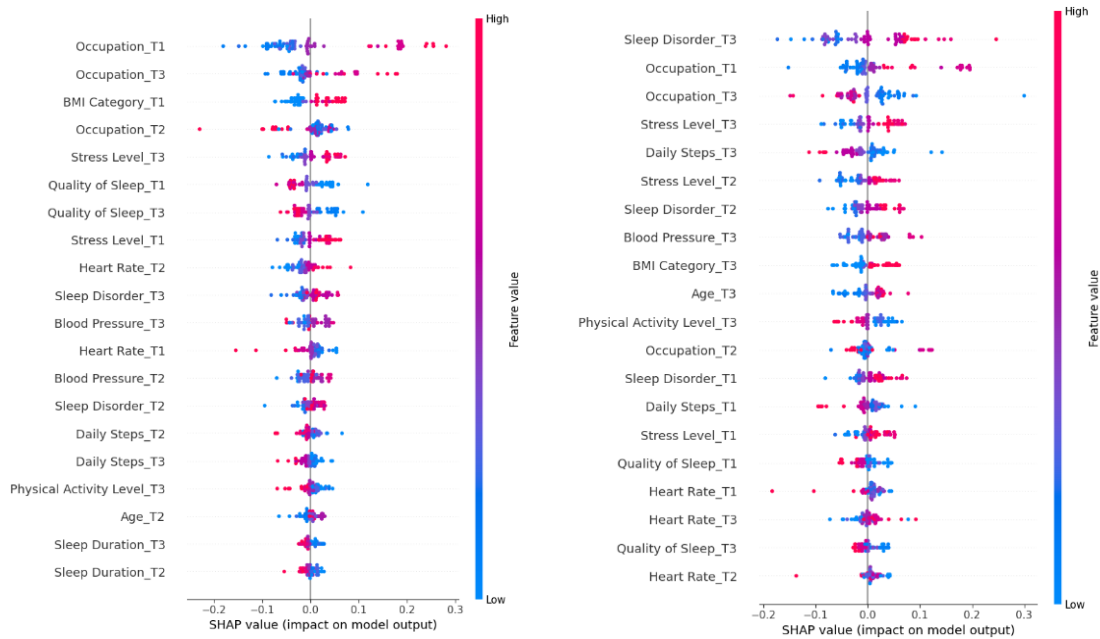

*Fig. 2 SHAP Value on LSTM and TCN*

The experiments aimed to evaluate the effectiveness and interpretability of LSTM and TCN models for predicting sleep disorders using SHAP values. SHAP values explain the impact of each feature on the model's predictions, enhancing transparency and trustworthiness. The dataset included features related to individuals' health and lifestyle, such as age, occupation, sleep duration, quality of sleep, physical activity level, stress level, BMI category, blood pressure, heart rate, daily steps, and the presence of a sleep disorder.

The SHAP summary plots for both LSTM and TCN models, presented in Figures 2 (right and left), illustrate the impact of each feature at different time steps on the model's predictions. Features with higher SHAP values have a more significant influence on the predictions. The SHAP summary plots reveal the most influential features in predicting sleep disorders. For both LSTM and TCN models, features such as occupation, stress level, quality of sleep, and physical activity level were consistently significant across different time steps. Comparing the SHAP plots for LSTM and TCN models shows that both models capture similar patterns in feature importance. However, the TCN model tends to have slightly different distributions of SHAP values, indicating a difference in how these models learn temporal dependencies.

**Temporal Attention and Counterfactual Explanations**

In this study, we have incorporated Temporal Attention mechanisms and Counterfactual Explanations [15] to significantly enhance the interpretability of our model predictions. These methodologies not only aid in understanding the model's decision-making process but also contribute to more reliable and actionable insights.

We integrated a Temporal Attention mechanism within our LSTM model to refine the processing of sequential input data. This mechanism dynamically assigns weights to different time steps in the input sequence, allowing the model to prioritize information that is most predictive of the outcome.

The model selectively focuses on time steps that contain crucial information for prediction. This is particularly useful in complex time series data where not all moments are equally informative.

By visualizing which time steps the model attends to, we can better understand the factors driving the predictions. Figure 3 illustrates this process, depicting how attention scores vary across the sequence, highlighting periods of heightened importance.

This focused approach ensures that the model's predictions are based on relevant data, enhancing the reliability of its outcomes.

Counterfactual explanations provide actionable insights by illustrating how modifying specific input features can alter predictions. For instance, a scenario where reducing stress levels from 8 to 4 while maintaining a high quality of sleep score shifts the prediction from "disorder" to "no disorder" highlights the model's sensitivity to stress. Visualizing such changes through feature-impact plots demonstrates the critical thresholds influencing outcomes. These insights empower clinicians to devise targeted interventions, such as stress management strategies, to improve patient outcomes effectively.

This method is instrumental in identifying features that are critical to the model's decisions, offering a clear pathway to understanding and trusting AI judgments.

By modifying key features in a controlled manner, we observe how the prediction shifts, which helps in pinpointing the features that significantly influence the model's output.

These explanations are invaluable for practitioners who need to understand the model in a straightforward and practical manner. They enable healthcare professionals to explore hypothetical scenarios and better prepare for patient-specific conditions.

The combination of Temporal Attention and Counterfactual Explanations provides a comprehensive understanding of the model's behavior, enhancing trust and enabling healthcare professionals to make more informed decisions based on the model's outputs.

**Conclusion**

This study underscored the important role of explainable AI in the time-series analysis for predicting sleep disorders, utilizing LSTM and TCN models alongside SHAP values to enhance interpretability and offer actionable insights. The application of Temporal Attention mechanisms and Counterfactual Explanations has improved the understanding of critical time periods and the impact of input feature modifications, promoting more trustworthy AI-driven predictions in healthcare.

Our approach not only ensures reliable predictions but also fosters trust among healthcare professionals, which is essential for better patient outcomes. Looking ahead, we plan to refine our models' interpretability through advanced visualization techniques and expand the data spectrum by integrating diverse demographic and physiological data, including wearable device data and patient-reported outcomes. These enhancements, coupled with the development of real-time predictive systems for early interventions, aim to prevent severe sleep disorders more effectively. Continuously improving the robustness and explainability of our models remains a priority for advancing AI applications in healthcare.

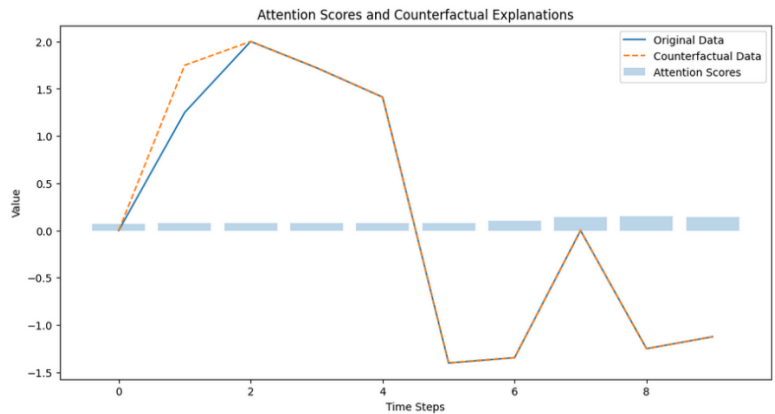

Fig. 3 Attention Scores and Counterfactual Explanations